\documentclass[runningheads]{llncs}
\usepackage[T1]{fontenc}
\usepackage{tikz}
\usepackage{amsmath}
\usepackage{graphicx}
\usepackage{amssymb}
\begin{document}
\title{Multi Robot Coordination in Highly Dynamic Environments: Tackling Asymmetric Obstacles and Limited Communication}
\titlerunning{Efficient Coordination in Dynamic Environments with Asymmetric Obstacles}
%
\author{Vincenzo Suriani\inst{1}\orcidID{0000-0003-1199-8358} \and
Daniele Affinita\inst{2}\orcidID{0009-0000-9347-9847} \and
Domenico D. Bloisi\inst{3}\orcidID{0000-0003-0339-8651} \and
Daniele Nardi\inst{2}\orcidID{0000-0001-6606-200X}}
\authorrunning{Suriani et al.}
%
\institute{Department of Engineering - University of Basilicata, Potenza (Italy)
\email{vincenzo.suriani@unibas.it}\\
\and
Department of Computer, Control, and Management Engineering ``Antonio Ruberti'', Sapienza University of Rome, Rome (Italy)\\
\email{affinita.1885790@studenti.uniroma1.it, nardi@diag.uniroma1.it}
\and 
Dept. of International Humanities and Social Sciences, 
International University of Rome, Rome (Italy),
\email{domenico.bloisi@unint.eu}
}
\maketitle              
\begin{abstract}
Coordinating a fully distributed multi-agent system (MAS) can be challenging when the communication channel has very limited capabilities in terms of sending rate and packet payload. When the MAS has to deal with active obstacles in a highly partially observable environment, the communication channel acquires considerable relevance. In this paper, we present an approach to deal with task assignments in extremely active scenarios, where tasks need to be frequently reallocated among the agents participating in the coordination process.
Inspired by market-based task assignments, we introduce a novel distributed coordination method to orchestrate autonomous agents' actions efficiently in low communication scenarios. In particular, our algorithm takes into account asymmetric obstacles. While in the real world, the majority of obstacles are asymmetric, they are usually treated as symmetric ones, thus limiting the applicability of existing methods. To summarize, the presented architecture is designed to tackle scenarios where the obstacles are active and asymmetric, the communication channel is poor and the environment is partially observable. Our approach has been validated in simulation and in the real world, using a team of NAO robots during official RoboCup competitions. Experimental results show a notable reduction in task overlaps in limited communication settings, with a decrease of 52\% in the most frequent reallocated task.

\end{abstract}

\section{Introduction}

\begin{figure}[ht]
    \centering
    \includegraphics[width=.95\columnwidth, trim={6em 0 6em 0}, clip]{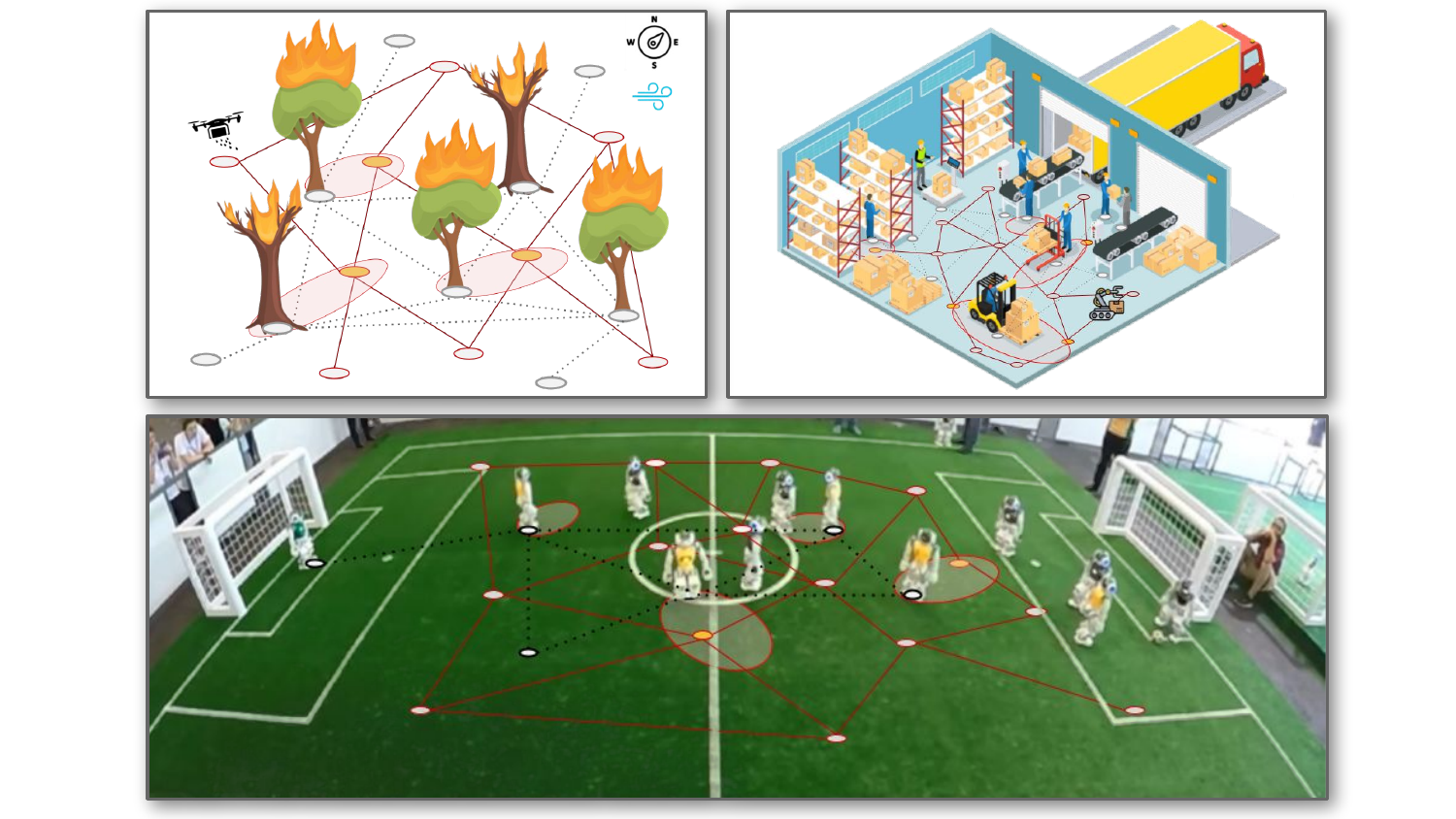}
    \caption{Various scenarios highlighting the critical role of representing obstacle dynamics in significantly improving both environmental modeling accuracy and overall team performance. Top-left: A wildfire situation necessitates agents to compute trajectories while accounting for wind and trees. Top-right: Autonomous agents in a warehouse environment must accurately represent human presence to facilitate effective cooperation. Bottom: A team of soccer robots adapts its coordination strategies based on the current formation of the opposing team.}
    \label{fig:intro}
\end{figure}

Research in autonomous robots and multi-robot systems with limited communication bandwidth is fundamental for creating realistic assumptions for lack of communication problems that are typically encountered in real world robotic applications. 
By developing efficient communication strategies and coordination algorithms for robots operating under constrained bandwidth, we can enable more robust and scalable applications in areas such as search and rescue, environmental monitoring, precision agriculture, and autonomous transportation. In fact, overcoming bandwidth limitations enhances the adaptability, reliability, and cost-effectiveness of multi-agent systems (MASs), promoting innovative solutions to complex problems when autonomous agent teams are deployed in practical scenarios.

State-of-the-art approaches for MASs operating in low bandwidth focus on optimizing communication protocols and control algorithms. These approaches often employ adaptive data compression and prioritization techniques to minimize the data exchanged between robots, reducing bandwidth requirements while maintaining essential information flow. 
However, current approaches have significant limitations, including reduced coordination efficiency, slower data exchange, increased latency, and heightened vulnerability to communication failures.

In this work, we present a coordination method that works efficiently in resource-constrained environments, ensuring robust and reliable performance even when communication bandwidth is limited. We do so by enhancing the overall situational awareness of the agents by providing them with a more refined understanding of the current state of the world, prediction capabilities, and better modeling of entities in the environment. In particular, we are interested in modeling MASs deployed in highly dynamic environments and operating under partial-observability, non-stationarity, and when both collaborative and adversarial entities are present.

As shown in Fig.~\ref{fig:intro}, our approach applies to diverse scenarios, significantly improving environmental modeling accuracy. We leverage \textit{asymmetric active obstacles} to better capture contextual dynamics. For instance, in forest fires, hazard zones shaped by wind require firefighters to adapt their movements accordingly. In warehouses, precise modeling of human and robot motions ensures efficiency and safety. In adversarial settings, agents must respond to opponents to optimize team performance. Across all these scenarios, accurate obstacle interpretation is essential for effective MAS decision-making.

The main contributions of this paper are:
\begin{enumerate}
    \item A market-based coordination system that limits performance-loss with low-network resources;
    \item A distributed world-model estimation scheme, which is updated locally by each agent in the team in order to propagate information even when no observation is received from teammates. Such a world-model estimator uses a Voronoi-based approach to apply corrections on the coordination systems upon task-assignment;
    \item A novel modeling of entities in the environment with an asymmetric entity-model (AM) that is used to integrate the concept of area-of-interest of an entity and propagate in time information of the operating environment more accurately.
\end{enumerate}

In order to validate our approach, we use both simulation and real data from a very challenging benchmark environment, i.e., the RoboCup, where a team of robots must collaborate effectively to compete in soccer matches. A RoboCup soccer match is suitable to showcase how our contribution can significantly reduce the number of failed task exchanges and improve the overall performance of a MAS operating under limited communication bandwidth.  

The remainder of the paper is organized as follows. Related work is analyzed in the next Section \ref{sec:relatedwork}, while our method is presented in Section \ref{sec:method}.  Experimental results are shown in Section \ref{sec:results}. Conclusions are drawn in Section \ref{sec:conclusions}.

\section{Related Work}
\label{sec:relatedwork}

Coordinating a team of autonomous agents is a challenging task, particularly in highly dynamic environments with strict constraints imposed by communication bandwidth limitations. Based on their level of robustness regarding communication failures and disturbances, existing approaches to agent coordination can be categorized into three classes: Perfect Communication, Low Loss, and Unreliable/No Communication.

\textbf{Perfect Communication.} This class includes methods supporting highly efficient coordination systems with stable communication. These typically assume no communication failures and omit world models. An early example is \cite{farinelli2005task}, which uses a token-passing mechanism for task allocation in an asynchronous distributed system. Lou et al. \cite{luo2011multi} propose an auction-based task allocation algorithm that groups tasks and respects precedence constraints. In \cite{weigel2002cs}, utility-based estimations guide task allocation in robot soccer, generating preferred positions for the team. MacAlpine et al. \cite{macalpine2013positioning} advance coordination via a formation system algorithm using a globally shared world model evaluated locally, with agents broadcasting their results after each evaluation. 

\textbf{Low Loss.} This group encompasses approaches capable of handling a limited number of data packages exchanged among the team of agents. Typically, they rely on the local estimation of a global world model to enable distributed decision-making \cite{vail2003multi}. These solutions estimate world states, mapping functions between agents and tasks, and strategic roles. Auction-based approaches are often used in these scenarios \cite{luo2011multi}, where the estimation of the mapping functions between agents and roles \cite{weigel2002cs}, or distributed world-states \cite{abeyruwan2012dynamic}, can alleviate communication losses. Moreover, more holistic approaches that can strategically change the task assignment of the team of agents have been published, where the agents exploit contextual knowledge for task negotiation \cite{7759298} and actively influence package broadcasting strategies \cite{rofer2022b}. In good and low-loss network conditions, consensus-based approaches have also been used\cite{consensus1}, \cite{consensus2}, but often, in very dynamic applications, this can lead to delayed task assignment procedures. 

\textbf{Unreliable/No Communication.} Algorithms in this class are designed to operate in extreme environments where agents share very little or no information among each other, but still need to guarantee a certain level of performance. Some approaches in this category leverage slowly changing features of the environment. For example, to model moving entities in an environment, promising solutions represent areas where such entities are likely to persist using Voronoi schemas \cite{OffensivePlacementVoronoi2014} \cite{rofer2022b}. Voronoi Diagrams (VDs) are a simple and effective way of modeling environments, representing entities at various levels of abstraction. In \cite{drones7100598}, a distributed control algorithm based on VD partitioning collision cones has been proposed to coordinate the navigation of a swarm of unmanned ground vehicles interacting with a localized human operator in unknown cluttered environments. Results show highly accurate non-rigid swarm motion capabilities for several navigation modes, all including collision and obstacle avoidance.  

To address limited communication range, in both obstacle-free and obstacle-rich environments, \cite{limited_comm_voronoi2019} introduced the centric multiplicatively weighted Voronoi configuration with a distributed coverage control strategy under which the agents' configuration is guaranteed to converge to this optimal configuration. Once a VD is built to model an environment, its nodes can carry different information that needs to be evaluated according to the current situation and different contexts (see Fig.~\ref{fig:intro} for examples). To model asymmetries of environmental entities effectively, VDs can be extended. In \cite{okabe2009spatial}, a Voronoi Diagram is generalized to a Line Voronoi Diagram (LVD), where sites are line segments, leading to lines and parabolic arcs. However, Voronoi cells might not be connected if the corresponding elements in the target set cross each other, leading to narrow areas not suitable for autonomous agents. A solution to this problem is presented in \cite{kropatsch2018line}, \cite{gabdulkhakova2018confocal}, where Elliptical Line Voronoi Diagrams (ELVD) are introduced, enabling longer line segments to have a greater influence on the Voronoi boundary thanks to a point-to-line segment computed using the Confocal Ellipse-based Distance.


Our method belongs to the class of Unreliable/No Communication approaches. We build on top of solutions proposed for low-bandwidth scenarios by leveraging distributed world knowledge and task-role assignments. However, we exploit non-centered elliptical Voronoi diagrams to support the coordination system of our team of agents when no data is received. We do so by applying corrections to the agent positioning and obstacles in the environment by updating the underlying Voronoi Diagram. We leverage the extended general concept of ELVDs, incorporating displacement to represent obstacles that have their own areas of interest. This integration is part of a market-based coordination system where the primary requirement is agent autonomy, allowing them to function effectively in situations with restricted communication.


\section{Proposed Method} 
\label{sec:method}
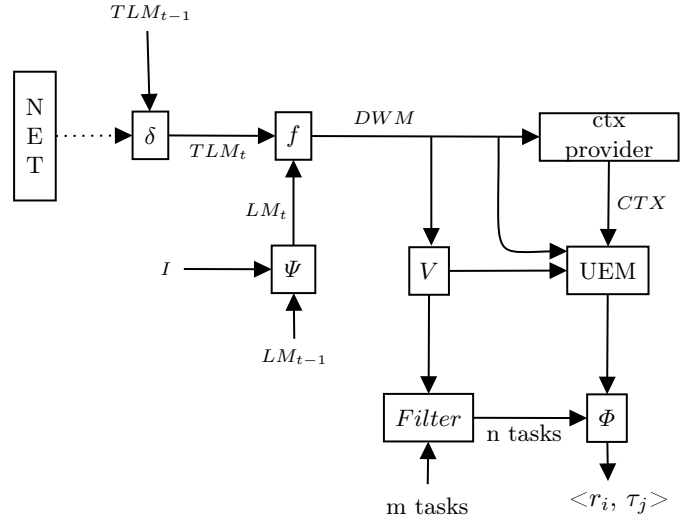
\begin{figure}[t]
    \centering
    \tikzset{every picture/.style={line width=0.75pt}} 

\begin{tikzpicture}[x=0.75pt,y=0.75pt,yscale=-1,xscale=1]

\draw    (445,205.7) .. controls (445.39,272.63) and (438.57,263.34) .. (477.25,263.32) ;
\draw [shift={(479.68,263.33)}, rotate = 180.55] [fill={rgb, 255:red, 0; green, 0; blue, 0 }  ][line width=0.08]  [draw opacity=0] (8.93,-4.29) -- (0,0) -- (8.93,4.29) -- cycle    ;
\draw    (411.07,205.15) -- (411.09,257.53) ;
\draw [shift={(411.09,260.53)}, rotate = 269.98] [fill={rgb, 255:red, 0; green, 0; blue, 0 }  ][line width=0.08]  [draw opacity=0] (8.93,-4.29) -- (0,0) -- (8.93,4.29) -- cycle    ;

\draw    (200.66,172.71) -- (221.66,172.71) -- (221.66,237.71) -- (200.66,237.71) -- cycle  ;
\draw (211.16,205.21) node   [align=left] {\begin{minipage}[lt]{11.78pt}\setlength\topsep{0pt}
\begin{center}
N
E
T 
\end{center}

\end{minipage}};
\draw    (260.41,192.83) -- (278.41,192.83) -- (278.41,216.83) -- (260.41,216.83) -- cycle  ;
\draw (269.41,204.83) node    {$\delta $};
\draw (247.58,137.59) node [anchor=north west][inner sep=0.75pt]  [font=\scriptsize]  {$TLM_{t-1}$};
\draw    (332.56,193.15) -- (350.56,193.15) -- (350.56,217.15) -- (332.56,217.15) -- cycle  ;
\draw (341.56,205.15) node    {$f$};
\draw (287.43,207.71) node [anchor=north west][inner sep=0.75pt]  [font=\scriptsize]  {$TLM_{t}$};
\draw (277.27,272.26) node  [font=\scriptsize]  {$I$};
\draw    (330.57,260.45) -- (352.57,260.45) -- (352.57,284.45) -- (330.57,284.45) -- cycle  ;
\draw (341.57,272.45) node    {$\Psi $};
\draw (342.06,317.02) node  [font=\scriptsize]  {$LM_{t-1}$};
\draw (328.06,243.02) node  [font=\scriptsize]  {$LM_{t}$};
\draw    (465.78,193.67) -- (535.78,193.67) -- (535.78,217.67) -- (465.78,217.67) -- cycle  ;
\draw (500.78,205.67) node   [align=left] {\begin{minipage}[lt]{44.93pt}\setlength\topsep{0pt}
\begin{center}
{\footnotesize ctx provider}
\end{center}

\end{minipage}};
\draw (370.1,191.04) node [anchor=north west][inner sep=0.75pt]  [font=\scriptsize]  {$DWM$};
\draw    (479.62,261) -- (520.62,261) -- (520.62,285) -- (479.62,285) -- cycle  ;
\draw (500.12,273) node   [align=left] {\begin{minipage}[lt]{25.4pt}\setlength\topsep{0pt}
\begin{center}
UEM
\end{center}

\end{minipage}};
\draw    (399.96,261.29) -- (419.96,261.29) -- (419.96,285.29) -- (399.96,285.29) -- cycle  ;
\draw (409.96,273.29) node    {$V$};
\draw (503.1,234.04) node [anchor=north west][inner sep=0.75pt]  [font=\scriptsize]  {$CTX$};
\draw    (489.67,335.56) -- (509.67,335.56) -- (509.67,359.56) -- (489.67,359.56) -- cycle  ;
\draw (499.67,347.56) node    {$\Phi $};
\draw    (387.17,335.33) -- (432.17,335.33) -- (432.17,359.33) -- (387.17,359.33) -- cycle  ;
\draw (409.67,347.33) node    {$Filter$};
\draw (409.14,391.89) node  [font=\small] [align=left] {m tasks};
\draw (457.93,354.94) node  [font=\normalsize] [align=left] {{\small n tasks}};
\draw (478.86,383.34) node [anchor=north west][inner sep=0.75pt]  [font=\normalsize] [align=left] {<$r_i$, $\tau_j$>};
\draw  [dash pattern={on 0.84pt off 2.51pt}]  (221.66,205.14) -- (257.41,204.91) ;
\draw [shift={(260.41,204.89)}, rotate = 179.63] [fill={rgb, 255:red, 0; green, 0; blue, 0 }  ][line width=0.08]  [draw opacity=0] (8.93,-4.29) -- (0,0) -- (8.93,4.29) -- cycle    ;
\draw    (267.86,152.19) -- (268.97,189.83) ;
\draw [shift={(269.06,192.83)}, rotate = 268.31] [fill={rgb, 255:red, 0; green, 0; blue, 0 }  ][line width=0.08]  [draw opacity=0] (8.93,-4.29) -- (0,0) -- (8.93,4.29) -- cycle    ;
\draw    (278.41,204.87) -- (329.56,205.1) ;
\draw [shift={(332.56,205.11)}, rotate = 180.26] [fill={rgb, 255:red, 0; green, 0; blue, 0 }  ][line width=0.08]  [draw opacity=0] (8.93,-4.29) -- (0,0) -- (8.93,4.29) -- cycle    ;
\draw    (286.27,272.29) -- (327.57,272.41) ;
\draw [shift={(330.57,272.42)}, rotate = 180.16] [fill={rgb, 255:red, 0; green, 0; blue, 0 }  ][line width=0.08]  [draw opacity=0] (8.93,-4.29) -- (0,0) -- (8.93,4.29) -- cycle    ;
\draw    (341.57,260.45) -- (341.57,220.15) ;
\draw [shift={(341.57,217.15)}, rotate = 89.99] [fill={rgb, 255:red, 0; green, 0; blue, 0 }  ][line width=0.08]  [draw opacity=0] (8.93,-4.29) -- (0,0) -- (8.93,4.29) -- cycle    ;
\draw    (341.96,307.52) -- (341.74,287.45) ;
\draw [shift={(341.7,284.45)}, rotate = 89.38] [fill={rgb, 255:red, 0; green, 0; blue, 0 }  ][line width=0.08]  [draw opacity=0] (8.93,-4.29) -- (0,0) -- (8.93,4.29) -- cycle    ;
\draw    (350.56,205.18) -- (462.78,205.54) ;
\draw [shift={(465.78,205.55)}, rotate = 180.18] [fill={rgb, 255:red, 0; green, 0; blue, 0 }  ][line width=0.08]  [draw opacity=0] (8.93,-4.29) -- (0,0) -- (8.93,4.29) -- cycle    ;
\draw    (500.66,217.67) -- (500.27,258) ;
\draw [shift={(500.24,261)}, rotate = 270.57] [fill={rgb, 255:red, 0; green, 0; blue, 0 }  ][line width=0.08]  [draw opacity=0] (8.93,-4.29) -- (0,0) -- (8.93,4.29) -- cycle    ;
\draw    (500.05,285) -- (499.76,332.56) ;
\draw [shift={(499.75,335.56)}, rotate = 270.34] [fill={rgb, 255:red, 0; green, 0; blue, 0 }  ][line width=0.08]  [draw opacity=0] (8.93,-4.29) -- (0,0) -- (8.93,4.29) -- cycle    ;
\draw    (409.27,380.89) -- (409.49,362.33) ;
\draw [shift={(409.53,359.33)}, rotate = 90.69] [fill={rgb, 255:red, 0; green, 0; blue, 0 }  ][line width=0.08]  [draw opacity=0] (8.93,-4.29) -- (0,0) -- (8.93,4.29) -- cycle    ;
\draw    (409.92,285.29) -- (409.73,332.33) ;
\draw [shift={(409.72,335.33)}, rotate = 270.22] [fill={rgb, 255:red, 0; green, 0; blue, 0 }  ][line width=0.08]  [draw opacity=0] (8.93,-4.29) -- (0,0) -- (8.93,4.29) -- cycle    ;
\draw    (432.17,347.39) -- (486.67,347.52) ;
\draw [shift={(489.67,347.53)}, rotate = 180.14] [fill={rgb, 255:red, 0; green, 0; blue, 0 }  ][line width=0.08]  [draw opacity=0] (8.93,-4.29) -- (0,0) -- (8.93,4.29) -- cycle    ;
\draw    (499.86,359.56) -- (500.12,376.34) ;
\draw [shift={(500.17,379.34)}, rotate = 269.11] [fill={rgb, 255:red, 0; green, 0; blue, 0 }  ][line width=0.08]  [draw opacity=0] (8.93,-4.29) -- (0,0) -- (8.93,4.29) -- cycle    ;
\draw    (419.96,273.26) -- (476.62,273.08) ;
\draw [shift={(479.62,273.07)}, rotate = 179.82] [fill={rgb, 255:red, 0; green, 0; blue, 0 }  ][line width=0.08]  [draw opacity=0] (8.93,-4.29) -- (0,0) -- (8.93,4.29) -- cycle    ;

\end{tikzpicture}  
    \caption{The overall architecture of \emph{DWM} and \emph{DTA}. The input is represented by a network event. When events are missing, prediction models extend prior estimations. Local models merge into DWM, which is used to select valuable contexts and assign utility values to <robot, task> pairs. An optimal configuration (V) aligns roles with available robots. Finally, roles are assigned to maximize cumulative utilities.}
    \label{fig:flowchart}
\end{figure}

We propose a market-based, distributed approach for multi-agent coordination that is able to deal with a lack of information for an extended period of time. This specific topic has been overlooked in previous work about coordination, which mainly focuses on a standard situation in which it is always possible to share information among the agents. However, in a real-world application, it may happen that robot communication is not working or is delayed, especially when the communication medium is the network. Our method focuses on addressing high communication loss situations, leveraging the prediction models to compensate for the limited information exchange among agents.
In particular, to improve the predictive model, we include the assumption that each obstacle in the real world has an area of interest. This area can be favorable or unfavorable for an agent, but it comes from the environment modeling.

Fig. \ref{fig:flowchart} shows the overall architecture of our system. It is made of different modules, aiming to guarantee the execution in an environment where the agent may encounter unpredictable situations. The main components are: \emph{a distributed world modeling, a position provider based on the Voronoi diagram, and a distributed task assignment procedure}.

In order to represent the operative scenario, we consider $M$ tasks, denoted as $T = \{\tau_1, \hdots, \tau_m\}$, and $N$ robots, denoted as $R = \{r_1, \hdots, r_n\}$, where in general $M > N$. Furthermore, we assume that we possess knowledge of an optimal robot placement configuration depending on the world state. In our study, a central theme 
is the execution of both task assignment and world modeling in a distributed manner, without exchanging further information.

It is worth mentioning that, in the distributed world modeling procedure, the obstacle representation procedure
\begin{enumerate}
    \item takes into account an area of interest for each obstacle;
    \item computes an ellipse, which is used for defining the desired configuration of the MAS.
\end{enumerate}
This representation allows for a better propagation of the agents' local models in the absence of network information or new observations. 

The distributed world modeling is achieved by fusing the information from all the robots, updated by a transition model that each robot adopts to keep a coherent representation of the world even under low packet rates circumstances. To easily propagate the information coming from the robot, the set of desirable positions is initially chosen using a Voronoi-based position generator. At the end of the procedure, each robot is capable of self-assigning a role and being aware of the teammates' roles without an explicit information exchange.


\subsection{Distributed World Model}

A prerequisite for an effective distributed task assignment (DTA) algorithm is to have an accurate representation of the world. The inputs that affect the models are the robot's perceptions referred to as $I$ and the events $e$ sent from other robots through a common network. We refer to the local model of the robot itself at time $t$ as $LM_t$, representing the robot’s instantaneous belief about the world. Moreover, each robot maintains a local model at time $t$ for each teammate $j$, denoted as $TLM_{j,t}$. This represents the belief that the robot maintains about what each teammate believes about the world.

We categorize events as representations of key situations arising within the robot's operational environment. For instance, in the context of a robot soccer match, an event may be triggered when a robot detects the referee’s whistle. Another example occurs when no team member has perceived the ball for a certain period; if a robot subsequently locates the ball, the situational context shifts, prompting an event to inform the other agents.
It is important to note that these events are generally not generated at fixed intervals but rather occur asynchronously, reflecting the inherently dynamic nature of real-world scenarios. As a result, there may be periods during which no events are transmitted across the network.

To address this limitation, we employ a function denoted as $\delta$ to update the local model of teammate agents, referred to as $TLM_j$ for robot j. Specifically, this function merges the $TLM$ of the previous step with any received event, when available. In the absence of a received event, we uses a predictive function to compute a probabilistic model of the world. In particular, we rely heavily on Kalman Filters, where a prediction is made when no updates are available, and the model is corrected with observations when updates arrive. For instance, Kalman Filters are used to propagate the ball’s position using the last measured velocity and updating it according to a ball-rolling physics model. Moreover, we use a particle filter for robot localization, where dynamic updates are performed using odometry, and the localization is refined when the robot observes a landmark. This approach enables us to avoid sending the whole Local Model through the network, thus obtaining a good estimation using only the available information. 
\begin{equation} 
    TLM_{j,t} = \delta(TLM_{j,t-1}, e)
\end{equation}
A function $\Psi$ updates the agent's local model $LM$ by incorporating input data received from the sensors. In the absence of communications, we can iterate the process, predicting the current state of data computed by modeling modules in the previous frame using the predictive models discussed before.
\begin{equation} 
    LM_t = \Psi(LM_{t-1}, I)
\end{equation}
Having an updated version of the Local Model of every agent $\{TLM_{1,t},\\ \ldots, TLM_{n-1,t}, LM_t\}$, it is possible to reconstruct the \emph{Distributed World Model} ($DWM$) through the utilization of a merging function, represented by $f$, which combines these individual local models. This merging function varies depending on the component being merged. In particular, local obstacles are merged with global obstacles by clustering them using DBSCAN \cite{9356727}. For the ball position, outliers are removed, and a confidence-weighted average of the local and global positions is computed.
\begin{equation} 
    DWM_t = f(TLM_{1,t}, \hdots, TLM_{n-1,t}, LM_t)
\end{equation}
\subsection{Distributed Task Assignment}
In our study, our primary objective is to enhance team coordination and strategic decision-making by adapting to the evolving configurations of the world. 

The information condensed in the \emph{Distributed World Model} is used as input to function $V$ that generates a set of desirable positions representing the optimal agent configuration at that moment. Notice that the configuration generated is role-independent, and each point within it does not represent an assignment to a specific agent, but rather represents a collection of potential waypoints. The points generated from $V$ have two purposes:
\begin{enumerate}
    \item filter $N$ out of $M$ tasks;
    \item refine the Utility Estimation Matrix ($UEM$).
\end{enumerate}

The UEM represents the main data structure used to take into account the information of the other agents and simulate a task auction locally. It is composed of $N$ rows for agents and $M$ columns for tasks, where the entry $(i,j)$ contains a non-negative number representing the utility. This utility quantity defines how well the agent $i$ can perform the task $j$. The final goal is to maximize the sum of all the assignments.

The computation of $UEM$ is also influenced by the context selected from the context provider, allowing for adaptive role assignments based on the chosen strategy. The columns of the matrix are filtered using a module that compares the target of the roles with the waypoints derived from $V$. This filtering process transforms the matrix into an $N \times N$ square matrix, with the number of roles equal to the number of agents. Finally, a function $\Phi$ solves the optimization problem that maximizes the cumulative utility, providing the pairs $<r_i, \tau_j>$ starting from the filtered $UEM$. 
\begin{equation}
    \Phi(UEM, tasks) \xrightarrow{} <r_i, \tau_j> \; \forall i,j
\end{equation}
To assign tasks, we employ a simplified version of the Hungarian Algorithm, where roles are assumed to have priority. Specifically, we assume that the roles in the matrix are ordered by importance, meaning that a role at position $i$ has higher priority than a role at position $j$ if $i < j$. We initiate the assignment process by starting with the role in position 0 and assigning it to the agent associated with the row that maximizes its utility. Subsequently, we proceed to the next role in order of priority while considering the unassigned agents. This process allows each robot to simulate the potential assignments of other agents.
As the Distributed World Model is probabilistically identical for all agents, each agent will reach an identical set of assignments.
\subsection{Correction using the Voronoi Diagram and the Asymmetrical Obstacles} 

The $DWM_t$ presented improves with the information received from the environment. On the other hand, when the information exchange is poor, the adherence of the $DWM_t$ with the environment characteristic can decrease. This mainly leads to overlaps in the task assignment procedure and makes the distributed algorithms unstable. In this evolution of the proposed method, we exploit the potential of the $V$ function, by modeling the obstacles as Asymmetrical Entities, leading to an Elliptical Line Voronoi Representation that is involved also in the $UEM$ computation. 

The function $V$ is a domain-specific optimal function that we assumed a priori, and which is used in the selection of the best $N$ tasks and for the refinement of the UEM. The selection of the function is based on some precise aspects that are desired to maximize or minimize, according to the environment. 
In the aim to make the evolution of the multi-agent system stable, we need to make them in a configuration that tends to keep the agent's poses as differentiated as possible. To this end, the Voronoi Diagram is effective as part of the pose generation system. In the presented work, we adopted and tested two Voronoi-based systems, one using the Point Voronoi Diagram as a component to filter the desired positions of the agents, and another to take a further step in the modeling process, deploying an evolution of the Elliptical Line Voronoi Diagram as a mean to better capture the structure of the environment.
\begin{figure}[t]
    \centering
    \includegraphics[width=.90\columnwidth]{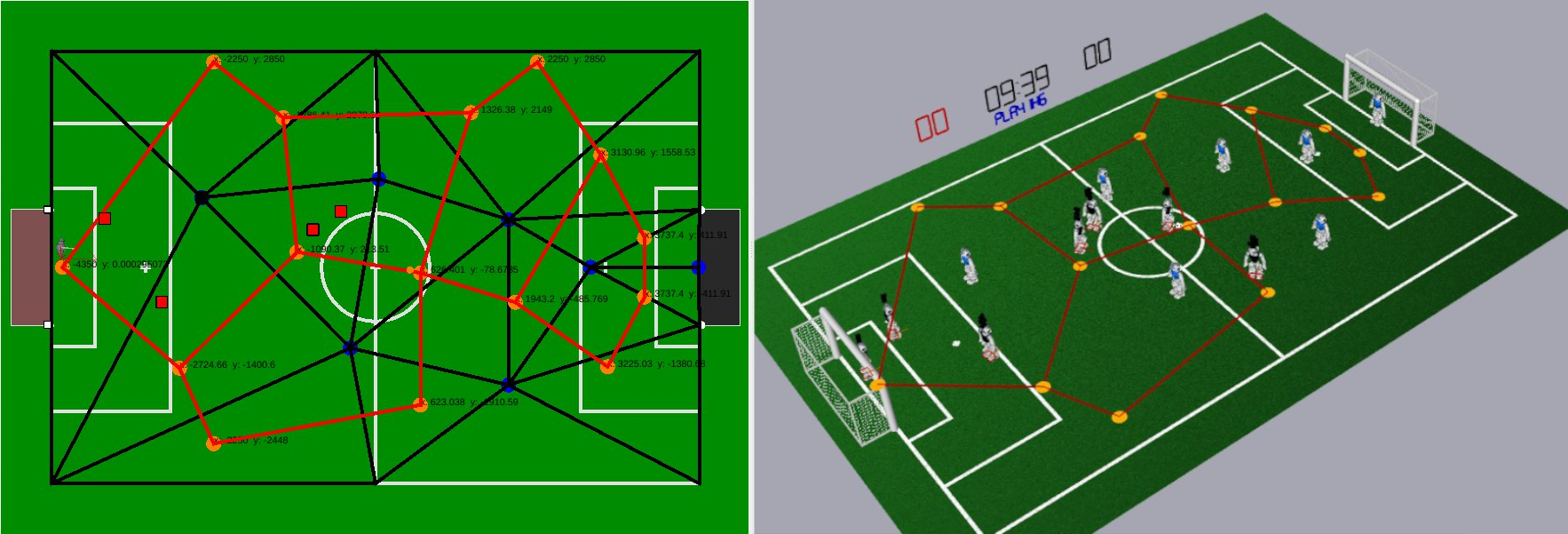}
    \caption{Voronoi graph in the event-based with VD schema approach, in 2D and 3D field view. In the 2D view in the left image, the blue points represent the opponent robots and the black connections depict the Delaunay Triangulation, while the red points are the Voronoi nodes and the red links are the Voronoi edges.
    In the 3D view on the right image, just the Voronoi nodes and edges are shown.}
    \label{fig:voronoi}
\end{figure}
\begin{figure}[t]
    \centering
    \includegraphics[width=.90\columnwidth]{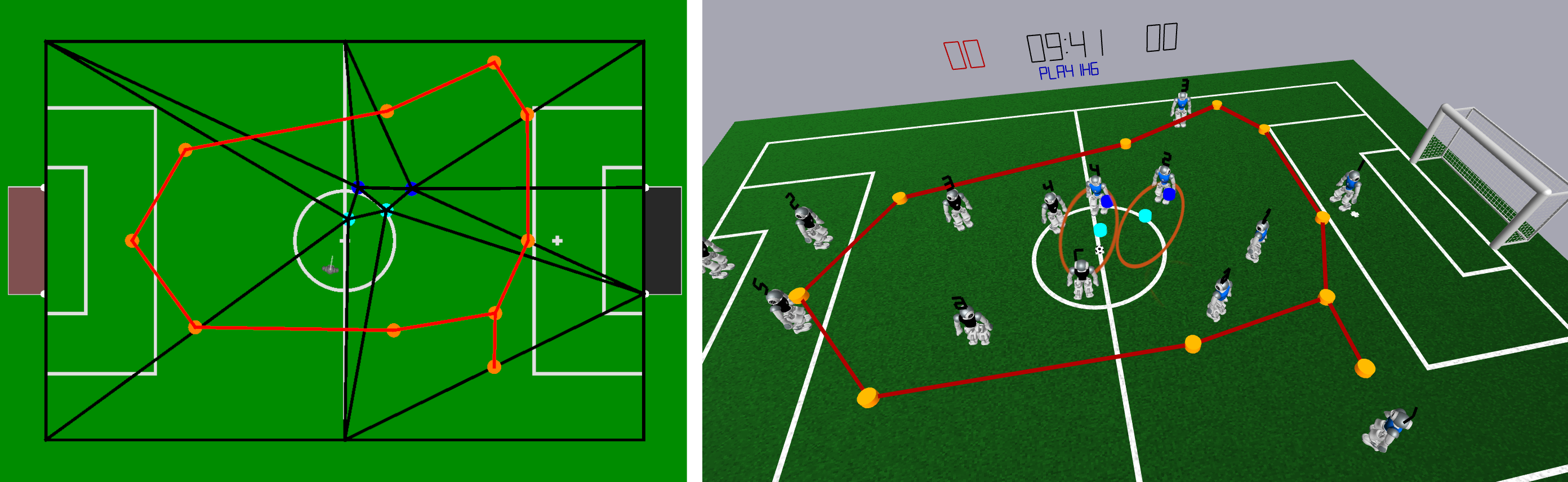}
    \caption{Voronoi graph in the event-based with ELVD schema approach, in 2D and 3D field view. The blue and cyan points represent the foci of the ellipse, modeling the asymmetric obstacle}
    \label{fig:voronoi_elliptical}
\end{figure}
\subsubsection{Point Voronoi Diagram}
\label{PVD}
Given a set of $n$ points in the plane (called sites), the Voronoi diagram is the partition of the plane in polygons based on the distance to them. In particular, it ensures every point inside the same region is closer to its associated site than to the others. 
Formally, define a metric distance $d$, we call $S=\{s_i| i=1,...,n\}$ the set of sites and $R=\{R_i|i=1,...,n \}$ the set of Voronoi regions, each one associated to the site $s_i$. Thus, taken a point $p$ of the plane:
\begin{equation}
    p \in R_i \iff d(p, s_i) \le d(p, s_j) \text{     } \forall j \ne i
\end{equation}
%
Every point $e$ such that $e \in R_i \land e \in R_j$ composes the Voronoi edge $E_{ij}$ between the polygons $R_i$ and $R_j$. So, the edge $E_{ij}$ is constituted by all the points that have the same distances with the sites $s_i$ and $s_j$, i.e.:
\begin{equation}
    E_{ij} = \{e | d(e, s_i) = d(e, s_j)\} \text{ with } e \in R_i \land e \in R_j
    \label{eq:VD_points}
\end{equation}
Every point $v$ that belongs to at least three different Voronoi regions is called a Voronoi node:
\begin{equation}
    v = R_i \cap R_j \cap R_k \cap ... \cap R_n
\end{equation}
%
In our case study, among all possible methods to build the graph, we first consider and construct it as the dual graph of the Delaunay triangulation, where the set of starting points is composed of all positions of active and passive obstacles, as depicted in Fig. \ref{fig:voronoi}. The final Voronoi nodes and edges represent respectively the furthest points from the rivals and the optimal path to follow between two adjacent nodes. In other words, Voronoi nodes constitute the optimal positions for the team disposal.
The filtering process for the N out of M tasks is done through the proximity of the tasks to the nodes. In this way, we can ensure to pick the most suitable tasks according to the dynamism of the environment. 
The refinement of the $UEM$ is performed by an offset applied to the tasks in the direction of their nearest nodes, which means the task positions approach the local optimal solution. 

\subsubsection{Elliptical Line Voronoi Diagram}
In the implementation that adopts classical VD, we are able to model the obstacle configurations and predict future states based on those. In order to improve the accuracy and to better propagate the local $DWM$, we modeled the obstacles with an Elliptical Line Voronoi Diagram (ELVD) with centroid offsets, capable of capturing the area of interest of the obstacles to better model and propagate them in the future steps. 
To this end, we have to consider extensions of the VD. The first modeling stage is to assign a direction of each obstacle, as a line. In the extension of the classical Line Voronoi diagram on the line set using
the Euclidean distance, the separating curves of the triangle are its angular bisectors which intersect at the center of the encircle. This leads to narrow passages not suitable for not punctiform agents. Hence, we adopted the ELVD, where the separating curve between the two line segments that share one endpoint is a hyperbolic branch\cite{gabdulkhakova2018confocal}. In the classical case, the sites $S=\{s_i| i=1,...,n\}$ are a finite set of points, and the metric used is the Euclidean distance; in the ELVD the original definition is extended, first, by considering a site to be a straight line segment and, second, by measuring the proximity of a point to the site using the parameters of a unique ellipse that passes through this point and takes the two endpoints of the line segment as its focal points. Differently from Euclidean distance in VD, proximity in the ELVD
is computed w.r.t. the Confocal Ellipse-based Distance, leading to define it as a set of points that have identical distance values for at least two sites. 
Starting from the definition of the ellipse E, we can have: 
\begin{equation}
    d(M, F_1)+d(M, F_2) = 2a
\end{equation}
in which the sum of the distances to two given points $F_0$ and $F_1$ is constant, given $a$ as the length of the semi-major axis of the ellipse. For the two focal points $F0$ and $F1$, we can have multiple ellipses, called Confocal Ellipses where each one is expressed by:
\begin{equation}
    E(a) = \{P \in{\mathbb{R}^2} | d(P, F0) + d(P, F1) = 2a\}, a \geq f
\end{equation}
where $f$ represents half the distance between $F_0$ and $F_1$. In our implementation, $f$ denotes the length of the area of interest of the obstacles and, from the Euclidean distance of the VD, the new distance between a point $P \in \mathbb{R}^2$  to the closest site $s$ becomes:
\begin{equation}
    d(P, s) = d(P, F_i) + d(P, F_{i+1}) - d(F_i, F_{i+1}) = 2(a - f)
\end{equation}
where the set of sites is represented by pairs of focal points:
\begin{equation}
    S = {(F_0, F_1), (F_2, F_3), ..., (F_{N-1}, F_N )}
\end{equation}
To model the obstacles we first place the $F_i$ on the obstacle. Then, given the direction of the obstacle's area of interest, we compute the position of $F_{i+1}$ using the length $f$. Then, we grow the ellipses to create the desired regions for the agents using the points that belong to at least three regions. This model allows for also describing active obstacles that evolve in their area of interest, allowing the agents to plan their moves with a model that better fits the environmental evolution even when no new observations are provided.

\subsubsection{Integration In the Coordination Process}
When the new $ELVD$ is computed, we update the coordination's utility vectors and the $UEM$ in order to blend the new desired position coming from the diagram with the task-related utilities. 

Finally, recalling the need to have a stable task allocation mechanism capable of facing a lack of network updates and very poor observations, we want to force the agents to minimize the chances of agents crossing unsuitable areas during a switch in the task allocation mechanism, from a coordination assignment to the next one.
To this end, we define a weighted correction to be assigned in the $UEM$ represented by:
\begin{equation}
W(v_i) = \sum_{o_j \in O} (1+cos(\theta_{i,j})) * e^{-\alpha \cdot \text{distance}}
\end{equation}
where $\theta_{ij}$ is the angle between the axis of the obstacle's ellipse main axis $o_j$ and point $v_i$, and $\text{distance}(o_j, v_i)$ is the Euclidean distance between $o_j$ and $v_i$.
\section{Experimental Results}
\label{sec:results}
In evaluating the viability of our proposed approach, we have highlighted various scenarios where implementing this approach can lead to an improved task assignment process. Among these scenarios, we opted to test its effectiveness in the challenging environment characterized by active obstacles, partial observability, and extremely limited communication: the RoboCup Standard Platform League (SPL)\footnote{https://spl.robocup.org/}. In fact, in this league, the current trend is to rely less on WiFi communication, in order to push the boundaries of the robot's capabilities in managing the distributed task assignment problem in challenging conditions. In particular, over the past few years, the network packet rate has been reduced from the original 5 packets per second per robot to a 1.200 total amount of packets per team per match. According to the rulebooks, from RoboCup 2019 to RoboCup 2022 the number of allowed packets per team has been reduced by 84\% \cite{rofer2022b}. In RoboCup 2023 the total number has been kept the same, but the number of playing robots per team increased from 5 to 7 (see Fig. \ref{fig:intro}). This further reduced the amount of packets per robot. Meanwhile, the size of the single packet has been reduced to half of its size (now it is only 128 B). This makes this scenario suitable for the deployment of such a coordination mechanism.

The approach has been used and qualitatively evaluated during RoboCup 2023\footnote{https://2023.robocup.org/en/robocup-2023/} and extensively evaluated in a simulated environment, SimRobot, released along with the B-Human code release\cite{roferb}. In the simulations, we emulated the sensors of the real robots and we further limited the perception capabilities by adding noise and false positives in the detection stage pipeline.

To assess the performance of our approach, we employed the metric of 'multiple role periods.' Specifically, we calculated, for each role, the total duration during a match in which more than two robots from the same team assumed that role simultaneously. Since the striker represents the most dynamic role with the highest priority, it best reflects coordination performance. 

We tested four different approaches, adding incrementally the presented features. In particular, we compared:
\begin{enumerate}
    \item \textbf{Fixed-rate coordination}: This experiment does not utilize events or $VD$ and it is the base implementation with fixed-rate network communication and no Voronoi prediction.
    \item \textbf{Event-based coordination without $\mathbf{\textit{VD}}$ schema}: This represents our previous approach, which serves as the baseline for improvement.
    \item \textbf{Event-based coordination with $\mathbf{\textit{VD}}$ schema}: This is the approach that uses the classical $VD$ to build a desired representation allowing for better predictions. An example of an execution is in Fig. \ref{fig:voronoi}.
    \item \textbf{Event-based coordination with $\mathbf{\textit{ELVD}}$  with asymmetric obstacles}: This is the last presented approach, that adds the Elliptical Line Voronoi Diagram with asymmetrical obstacles in order to represent the local world model capable of extending the robot's prediction capability for a longer time, as shown in Fig. \ref{fig:voronoi_elliptical}.
\end{enumerate}

\begin{figure}[t]
    \centering
    \includegraphics[width=0.8\columnwidth]{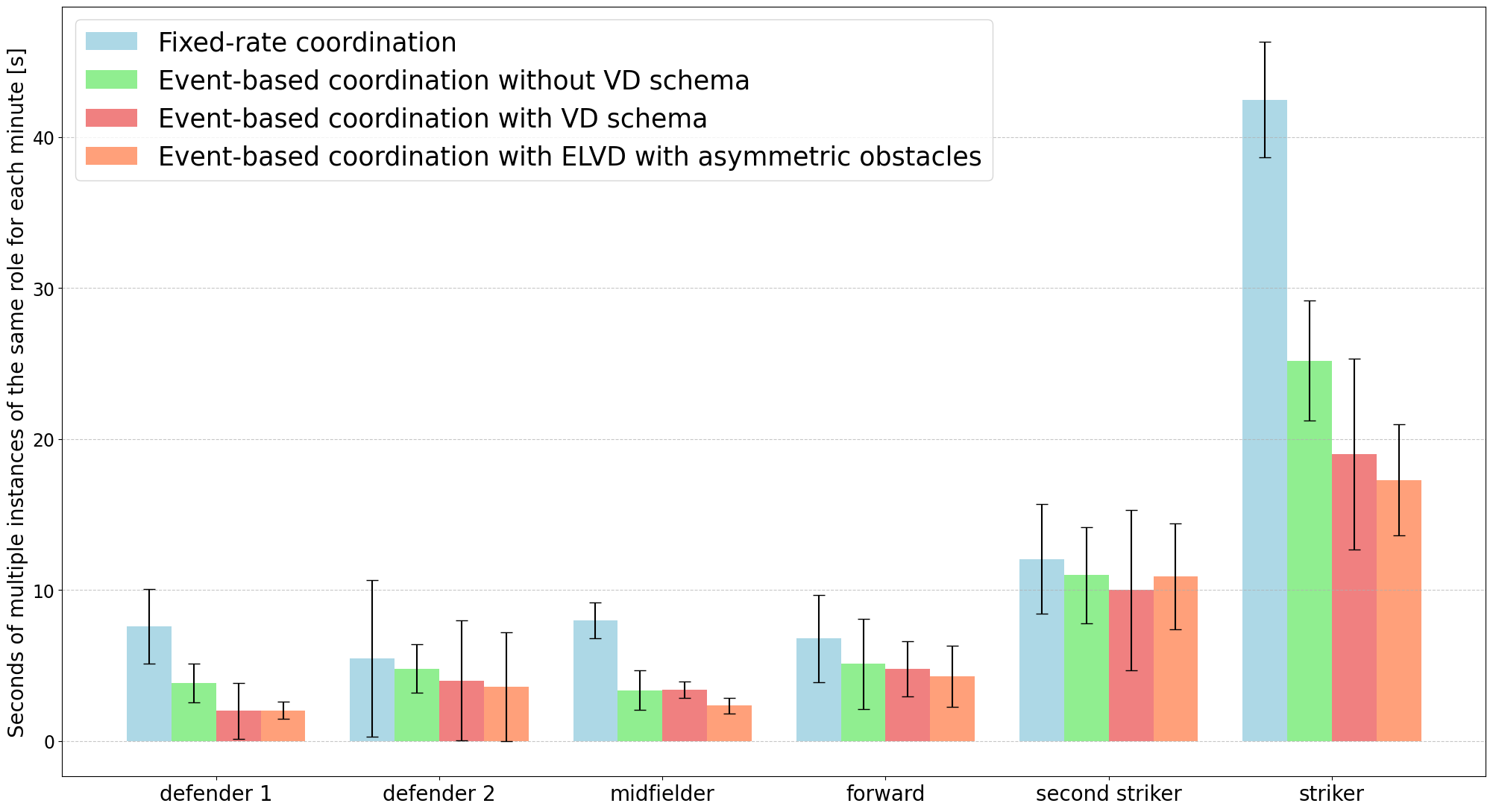}
    \caption{Role overlap over time: the x-axis represents the different roles, and the y-axis shows the rate of role overlaps per minute during a series of matches with 1200 total packets per team and 20 minutes of effective team play for each match. The majority of data has been collected in a simulated environment.}
    \label{fig:results}
\end{figure}

Fig.~\ref{fig:results} shows the cumulative role overlap duration. The x-axis represents the different roles, the y-axis shows the seconds of role overlaps per minute during a series of matches with 1200 total packets per team and 20 minutes of effective team play. As we can see from the results, the experiments demonstrate the improvement of the coordination stability during matches.
The improvement increases with the dynamism of the assigned tasks. In fact, the \textit{Striker} task is the ball holder behavior and it is assigned locally to the robot that is controlling the ball. Hence, from the \textit{Defender 1} to the \textit{Striker}, those tasks (and, hence, the roles) are typically ordered from far to close to the area at high dynamism. This leads to higher overlapping situations on the Striker and on the roles that used to fight for the ball and this is clearly visible on this plot. The main improvement derives from the use of events in the coordination mechanism but further improvements have been made possible only with a better modeling of the obstacle in order to propagate the local model in the absence of the external observations from the teammates.

It is worth noticing that, the use of our approach led to an improvement in terms of match score with respect to the baseline one. This is due to the strong reduction in overlapping situations, where robots fight for the same task thus becoming less efficient in the overall team play.

\section{Conclusions}
\label{sec:conclusions}


In this study, we have tackled the problem of coordinating a team of fully autonomous agents in a low communication setup. We have developed an innovative distributed coordination system based on market-based task assignments to deal with a possible reduction in network packet rates, which is an assumption that is often not considered in existing approaches to coordination. 

In particular, our method allows the agents to model the world locally, propagate world predictions when network data is limited, and consequently efficiently assign tasks to team members. We adopt a market-based approach in which every robot simulates an auction, locally assigning the available tasks to maximize the expected reward. We employ a Voronoi Graph to match the number of tasks with the number of available agents. Additionally, the Voronoi diagram is useful for calculating a portion of the reward, contributing to the differentiation of the total reward. To address limited communication, we use prediction models to compensate for missing information from other agents, sending messages only when specific events occur.

To validate our approach, we have considered the dynamic scenario of a RoboCup robot soccer match, where strict communication constraints hold.
Quantitative experiments, conducted both in simulation and on real data, demonstrate the effectiveness of the proposed method. In particular, the results clearly indicate that our approach effectively reduces task overlaps in low-communication scenarios, which is a critical factor in a multi-agent scenario.

\section*{Acknowledgments}
We acknowledge partial financial support from PNRR MUR project PE0000013-FAIR

%
%
 \bibliographystyle{splncs04}
 \bibliography{bibfile}
\end{document}